%% file: main.tex
\documentclass[10pt,twocolumn,letterpaper]{article}

\usepackage{cvpr}
\usepackage{times}
\usepackage{epsfig}
\usepackage{graphicx}
\usepackage{amsmath}
\usepackage{amssymb}
\usepackage{array}
\usepackage{url}

\usepackage{booktabs} 
\newcommand{\ra}[1]{\renewcommand{\arraystretch}{#1}}
\usepackage{multirow}
\usepackage[draft,inline,nomargin]{fixme}
\usepackage{soul}

\usepackage{cite}
\newcommand\relphantom[1]{\mathrel{\phantom{#1}}}

\DeclareMathOperator*{\argmin}{arg\,min}

\usepackage[breaklinks=true,bookmarks=false]{hyperref}

\cvprfinalcopy 

\def\cvprPaperID{2125} 

\ifcvprfinal\fi
\begin{document}

\title{Identifying First-person Camera Wearers in Third-person Videos}

\author{Chenyou Fan$^1$, Jangwon Lee$^1$, Mingze Xu$^1$,
Krishna Kumar Singh$^2$, Yong Jae Lee$^2$, 
\\ David J. Crandall$^1$ and Michael S. Ryoo$^1$ \\
$^1$Indiana University Bloomington\\
$^2$University of California, Davis\\
{\tt\small \{fan6,mryoo\}@indiana.edu}
}


\maketitle
\thispagestyle{empty}

\newcommand{\fp}{first-person\ }
\newcommand{\tp}{third-person\ }
\newcommand{\fandtp}{first- and third-person\ }

\begin{abstract}
We consider scenarios in which we wish to perform joint scene
understanding, object tracking, activity recognition, and other tasks
in environments in which multiple people are wearing body-worn
cameras while a \tp static camera also captures the scene.  To do
this, we need to establish person-level
correspondences across \fandtp videos, which is challenging because
the camera wearer is not visible from his/her own egocentric video,
preventing the use of direct feature matching. In this paper, we
propose a new semi-Siamese Convolutional Neural Network architecture
to address this novel challenge. We formulate the problem as learning
a joint embedding space for \fandtp videos that considers both
spatial- and motion-domain cues.  A new triplet loss function is
designed to minimize the distance between correct \fandtp matches
while maximizing the distance between incorrect ones. This end-to-end
approach performs significantly better than several baselines, in part
by learning the \fandtp features optimized for matching jointly with
the distance measure itself.
\end{abstract}

\input{intro}

\input{related}
\input{overview}

\input{network_structure}

\input{loss_function}

\input{experiment}

\input{case_study}

\input{conclusion}

\vspace{-3pt}
{\flushleft{
\textbf{Acknowledgements:}
This work was supported in part
by NSF (CAREER IIS-1253549) and 
the IU Office of the Vice Provost for Research, the College of Arts and Sciences, and the School of Informatics
and Computing through the Emerging Areas of Research Project ``Learning: Brains, Machines, and Children.''
CF was supported by a Paul Purdom Fellowship.
}}
{\small
\bibliographystyle{ieee}
\bibliography{egbib}
}

\end{document}

%% file: intro.tex
\section{Introduction}
Wearable cameras are becoming mainstream: GoPro and other first-person cameras
are used by consumers to record extreme sports and other activities, for example, while 
body-worn cameras are now standard equipment for many police and military
personnel~\cite{nytimes}. These cameras capture unique perspectives
that complement video data from traditional 
third-person cameras.  For instance, in a complex and highly dynamic
environment like a busy city street or a battlefield, third-person
cameras give a global view of the high-level appearance and events
in a scene, while \fp cameras capture 
ground-level evidence about  objects and people at a
much finer level of granularity.  The combination of video  from these
highly complementary views could be used to perform a variety of
vision tasks -- scene understanding, object tracking, activity
recognition, etc. -- with greater fidelity and detail than either
could alone.

\begin{figure}
\begin{center}
   \includegraphics[width=1.0\columnwidth]{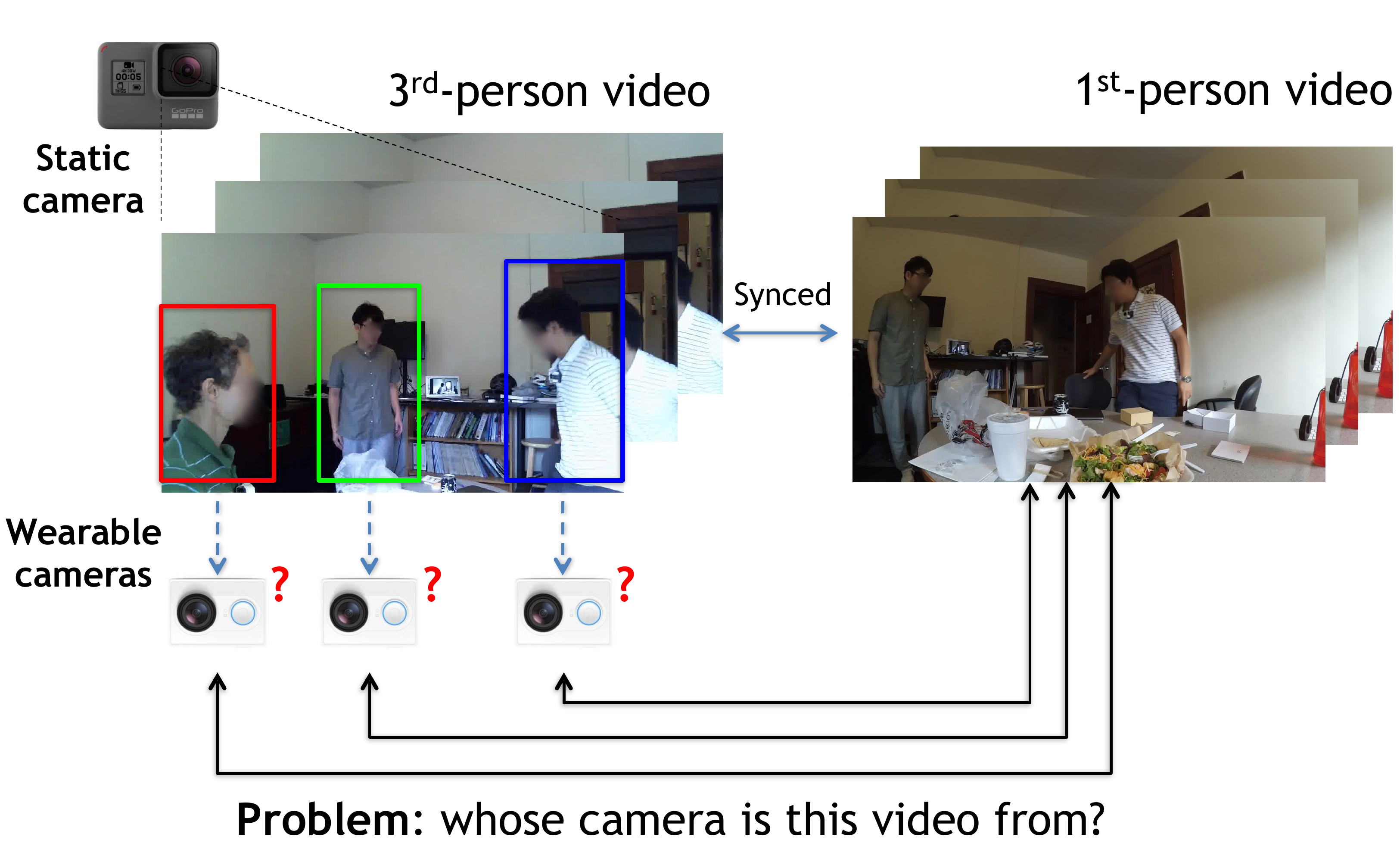}
\end{center}
\vspace{-12pt}
   \caption{One or more people wear first-person
cameras in a scene that is also recorded by a third-person camera.
We wish
to identify which person in the third-person view (left) was wearing the camera that
captured a first-person video (right). This is challenging because the camera fields of view
are very different and the camera wearer almost never appears in their own
first-person view.
   }
\label{fig:intro}		
\end{figure}

In these scenarios, multiple people may be in a
scene at any given time, with people regularly entering and exiting the view of the
\tp camera.  Some subset of these people may be wearing
\fp cameras, each of which is also capturing part of the
scene but from a highly dynamic point of view that changes as the
wearer moves.  Thus at any moment in time, some (possibly empty)
subset of people appear in any given camera's view, and each person
appears in some (possibly empty) subset of the \fandtp
cameras (and that person themselves may be wearing one of the
\fp cameras).  Compared to static cameras, \fp
video data is significantly more challenging because of camera motion, poor
scene composition, challenging illumination, etc.

Jointly solving computer vision problems across multiple \fandtp
cameras requires the crucial first step of establishing
correspondences between the people and the cameras, including (1)
identifying the same person appearing in different views, as well as
(2) matching a camera wearer in one view with their corresponding
\fp video.  The former problem is similar to person
identification and re-identification problems that have been studied
for \tp cameras~\cite{reidentificationbook}. These approaches typically rely
on matching visual and motion features of a person across different
views; the \fp camera version is similar in principle  but
significantly
more difficult due to the difference in perspectives and characteristics
of \fandtp video.

The second problem is even more challenging, since a person's
\textit{appearance} in one video may share few (if any) visual features
with his or her \textit{\fp visual field} of the same scene.  
For
instance,  a surveillance camera might capture a camera wearer
walking down the street, including her physical appearance, the 
cars parked on the street beside her, and the friends walking next to
her, while her front-facing \fp camera may capture \textit{none} of these 
because its field of view is down the street.
Finding correspondences thus cannot rely on direct appearance feature
matching. Instead, we must rely on  indirect sources of evidence
for finding correspondences: (i) matching \fp videos against an
estimate of what a person's field of view 
\textit{would look like} based on the \tp view of the scene; (ii)
matching estimates of a person's body movements based on the
camera motion of their \fp video to the movements observed by the \tp
static camera; and (iii) matching the (rare) moments when
part of a camera wearer's body or actions are directly visible in the
scene (e.g.\ when reaching for an object and both  \fandtp cameras
see the hand). See Figure~\ref{fig:intro}.

Despite its importance, we are aware of very little work that tries to
address this problem.  Several recent papers propose using multiple
cameras for joint \fp recognition~\cite{soran15,yonetani16,bambach15icmi,chan16hand}, but make
simplistic assumptions like that only one person appears in the scene.
Using visual SLAM to infer \fp camera trajectory and map to \tp
cameras (e.g., \cite{park-iccv2013,poleg2014head}) works well in some settings, but can 
fail for crowded environments when long-term precise localizations are needed and 
when \fp video has significant motion blur. 
Ardeshir and Borji~\cite{ardeshir16} match a set of egocentric
videos to people appearing in a top-view video using graph matching,
but assume that there are multiple \fp cameras sharing the same field
of view at any given time, and only consider  purely overhead \tp cameras
(not oblique or ground-level views).
We require a more general
approach that matches each individual \fp video with the corresponding
person appearing in an arbitrarily-positioned \tp camera.

In this paper, we present a new semi-Siamese Convolutional Neural
Network (CNN) framework to learn the distance metric between \fandtp
videos. The idea is to learn a joint embedding space between \fandtp
perspectives, enabling us to compute the similarity between
\textit{any given \fp video} and \textit{an individual human appearing
  in a \tp video.}  Our new semi-Siamese design allows for learning
low-level features specialized for \fp videos and for \tp videos
separately, while sharing higher-level representations and an
embedding space to permit a distance measure. Evidence from both scene
appearance and motion information is jointly considered in a novel
two-stream semi-Siamese CNN.  Finally, we introduce a new ``triplet''
loss function for our semi-Siamese network, and confirm its advantages
in our experiments on a realistic dataset.

%% file: related.tex
\section{Related work}

While many of the core problems and
challenges of recognition in \fp (egocentric) videos are shared with
traditional \tp tasks, \fp video tends to be much more challenging,
with highly dynamic camera motion and difficult imaging conditions.
Research has focused on extracting  features customized for
\fp video, including hand~\cite{lee14}, gaze~\cite{li13}, and
ego-motion cues~\cite{poleg14}. Other work has studied
object-based understanding for activity recognition~\cite{ramanan12},
video summarization~\cite{lee12,zheng12}, and recognition of
ego-actions~\cite{kitani11} and interactions~\cite{ryoo13}, 
but in single \fp videos.



Several recent papers have shown the potential for 
combining \fp video analysis with evidence from other types of synchronized video,
including from other \fp cameras~\cite{yonetani16,bambach15icmi}, multiple \tp cameras~\cite{soran15}, 
or even hand-mounted cameras~\cite{chan16hand}.
However, these papers assume
that a single person appears in each video, avoiding the
person-level correspondence problem.  Our work is complementary,
and could help generalize these approaches to 
scenarios in which multiple people appear
in a scene.

A conventional approach to our person correspondence problem might use
visual odometry and other camera localization techniques~\cite{slam,vo}
to estimate the 3-d trajectory of the wearable camera, which could
then be projected onto the static camera's coordinate system to identify the camera
wearer~\cite{park-iccv2013}.  However, this is problematic
in crowded or indoor environments where
accurate localization is difficult and people are standing close
together.  Precise online visual localization in indoor environments
with few landmarks is itself challenging, and not applicable when
cameras are not calibrated  or move too quickly and cause
motion blur.

 Perhaps the work most related to ours is that of Ardeshir
and Borji~\cite{ardeshir16}, which matches a set of egocentric
videos to a set of individuals in a top-view video using graph-based
analysis. This technique works well but
makes two  significant assumptions that limit its real-world
applicability. First, it requires the static camera to have a strictly
top-down (directly overhead) view, which is relatively uncommon in
the real world (e.g.\ 
wall-mounted surveillance cameras capture oblique views).  Second, it assumes that multiple
egocentric videos sharing the same field-of-view are available.  This
assumption is strong even if there are multiple people wearing cameras: the cameras may not share any field of
view due to relative pose or occlusions, and even if
multiple \fp videos with overlapping fields of view are recorded, some
users may choose not to share them due to privacy concerns,
for example.  
In contrast, we consider the more challenging
problem of matching each of multiple \fp cameras having arbitrary fields of
view with a static, arbitrarily-mounted \tp camera.

We believe this is the first paper to formulate \fandtp
video correspondence as an embedding space learning problem
and to present an end-to-end learning approach. Unlike previous
work~\cite{poleg2014head,yonetani15} which 
  uses hand-coded trajectory
features to match videos without any embedding learning, our method
is applicable in more complex environments (e.g.\ with
arbitrarily placed \fandtp cameras and arbitrary numbers of people).

%% file: overview.tex
\section{Our approach}
\label{sec:overview}

Given one or more \fp videos, our goal is to decide if each of the
people appearing in a \tp video is the wearer of one of the \fp
cameras.  The key idea is that despite having very different
characteristics, synchronized \fandtp videos are different
perspectives on the same general environment, and thus capture some of
the same people, objects, and background (albeit from two very
different perspectives).  This overlap may allow us to find similarities in
\textit{spatial-domain (visual) features}, while hopefully
ignoring differences due to perspective.  Meanwhile, corresponding
\fandtp videos are also two reflections of the same person performing
the same activity, which may allow us 
to find \textit{motion-domain feature} correspondences
between video types.





%

We formulate this problem in terms of learning embedding
spaces shared by \fandtp videos.  Ideally, these embeddings 
minimize the distance between the \fp video features 
observed by a camera wearer and the visual features of the same person
observed by a static \tp camera at the same moment, while
maximizing the distances between incorrect matches. We propose
a new semi-Siamese network architecture, detailed in the next section, to learn this embedding
space.
To handle the two modalities (motion and spatial-domain), we design a new two-stream Siamese
CNN architecture
where one stream captures temporal information using optical flow
(i.e., motion) and the other captures spatial information
(i.e., surrounding scene appearance), 
which we detail in Section~\ref{sec:twostream}.
We also consider two loss functions: a traditional contrastive loss
that considers pairs of samples, and a new triplet loss  that
 takes advantage of the fact that
both  positive 
and negative first-to-third-person 
pairings exist in the same scene. We describe these
losses in Section~\ref{sect:lossfunction}.


\subsection{Semi-siamese networks}

Our approach is based on Siamese networks with contrastive
loss functions, which enable end-to-end learning of both low-level visual features and
an embedding space (jointly optimizing them based on training data). The
original Siamese formulation~\cite{hadsell2006dimensionality} interprets the
network as a function 
$f(I; \theta)$ that maps each input video $I$ into an embedded point using parameters $\theta$, which are typically trained based on contrastive loss between embedding
of the positive and negative examples~\cite{bell15siggraph,schroff2015facenet}.
If we applied this approach to our problem, $I$ would be either the \fp or \tp video,
such that the network (i.e., function $f$
and parameters $\theta$) would be shared by both types of videos. 

However, \fandtp videos are very different,
even when recording the 
same event by the same person in the same location. We hypothesize 
that although the higher-level representations that capture
object- and action-level information in \fandtp videos might be
shared, the optimal low-level features (i.e., early convolutional filters)
may not be identical.

We thus propose a \textit{semi-Siamese} architecture to learn
the first- to third-person distance metric.  We find separate
parameters for \fandtp videos, which we call $\theta_1$ and $\theta_2$, respectively,
while forcing them to share a subset of parameters
$\theta$. Given a set $E$ of egocentric cameras and a set $P$ of
detected people in a \tp camera view, we can easily estimate the
person corresponding to a given egocentric camera $e \in E$ using this
embedding space,
%
%
%
\begin{equation}
p^*_e = \argmin_{p \in P} ||f(I_e; \theta_1, \theta) - f(I_p; \theta_2, \theta)||.
\end{equation}
%
%

%% file: network_structure.tex

\begin{figure*}[t]
\begin{center}
{\small{
\begin{tabular}{cc}
 \includegraphics[clip, trim=200 1150 200 100, width=3.4in ]{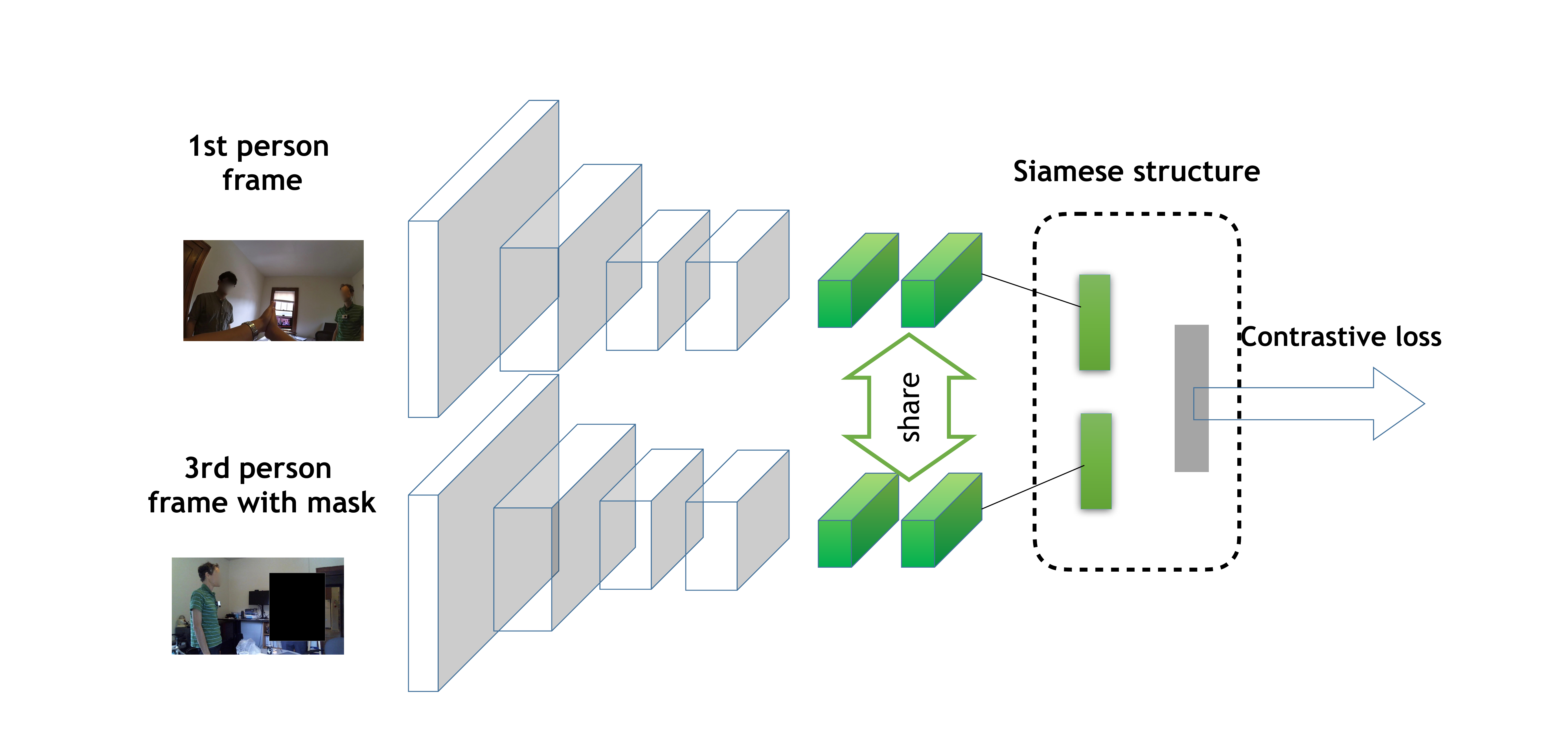} &
 \includegraphics[clip, trim=200 450 200 850, width=3.4in ]{image/spatio-temporal.pdf} \\
(a) Spatial-domain semi-Siamese network &
(b) Motion-domain semi-Siamese network \\ \\
\includegraphics[clip, trim=100 350 200 200, width=3.2in ]{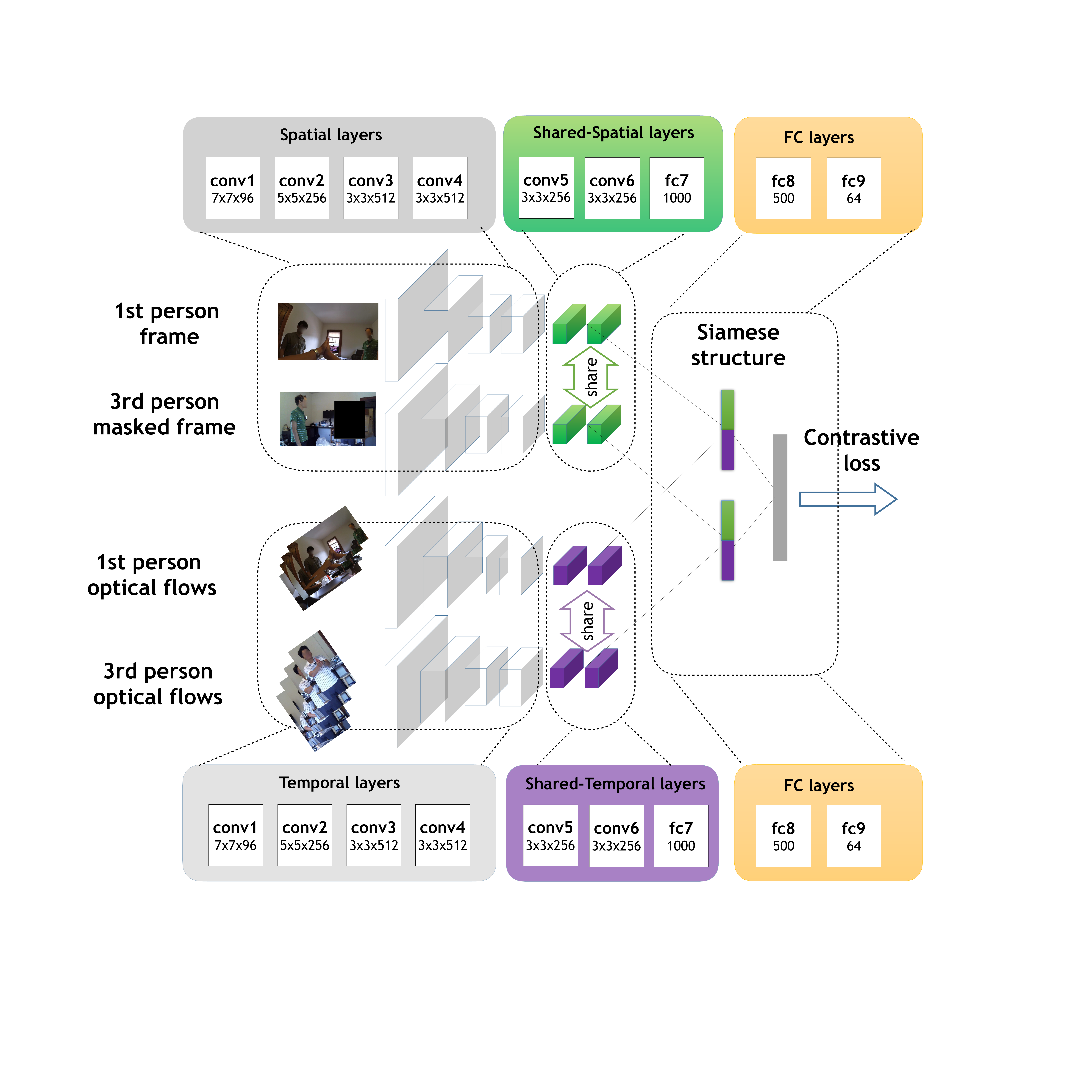} &
 \includegraphics[clip, trim=450 300 200 260, width=2.6in ]{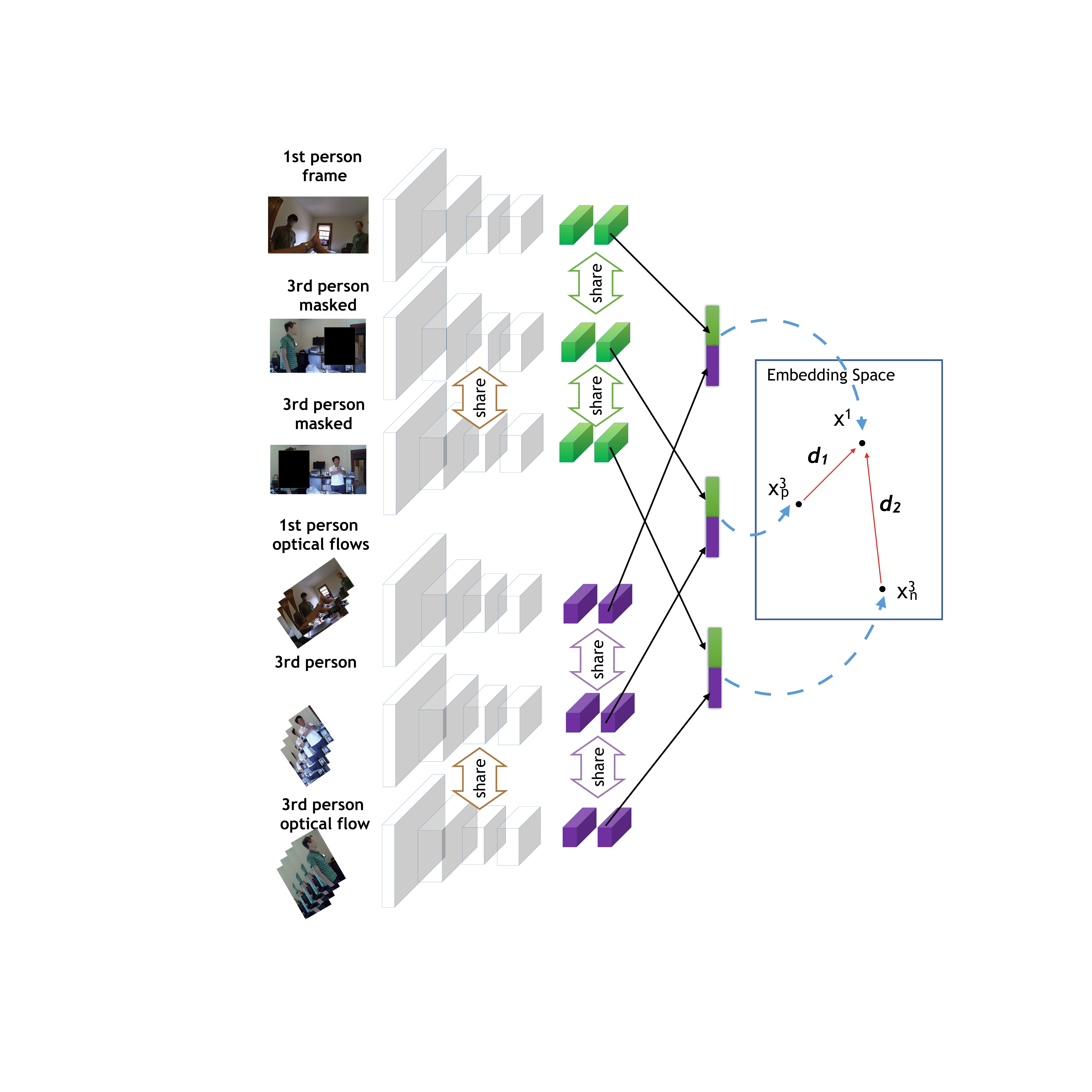} \\
(c) Two-stream semi-Siamese network &
(d) Two-stream semi-triplet network \\
\end{tabular}
}}
\end{center}
   \caption{\textit{Overview of our networks.} All networks receive
     features from time-synchronized \fandtp video frames, which
     during training consist of correct correspondences as positive
     examples and incorrect correspondences as negative
     examples. \textbf{(a) Spatial-domain network} is a semi-Siamese
     network with separate early convolutional layers (gray) and
     shared later layers (green). Corresponding input pairs consist of
     a \fp video frame and \tp frame with the corresponding person
     (camera wearer) masked out, since he or she is not visible in his
     or her own \fp camera.  \textbf{(b) Motion-domain network} is
     also semi-Siamese with a similar structure, except that it inputs
     stacked optical flow fields instead of images frames, and the \tp
     flow field consists of a crop of a single person.  \textbf{(c)
       Two-stream semi-Siamese network} combines both networks with a
     fully-connected layer that produces the final feature vector.
     \textbf{(d) Two-stream semi-Siamese network trained with triplet loss} receives three
     inputs during training: a \fp frame, the corresponding \tp frame
     with the correct person masked out, and the same \tp frame with a
     random incorrect person masked out.}
\label{fig:nets}
\end{figure*}


We now propose specific network architectures, first considering the two
feature modalities (spatial-domain and motion-domain) independently,
and then showing how to combine them into integrated two-stream networks.

\subsubsection{Spatial-domain network}

To learn the spatial-domain correspondences between \fandtp cameras,
our network receives a single frame of \fp video and the
corresponding frame of \tp video (Figure~\ref{fig:nets}(a)).  For the \tp video, we force
the network to consider one specific person by masking him or her out
from the rest of the frame, replacing their bounding box with
black pixels.  This is important because a camera wearer does not
appear in their own \fp video (with the occasional exception of
arms or hands). We thus encourage the network to learn a relationship
between a \fandtp video frame, with that person
\textit{removed from} the \tp scene.
%

As
shown in Figure~\ref{fig:nets}(a), each of the \fandtp branches maintains its own four
early convolution layers while sharing the last two convolution layers
and fully connected layers. The intuition here is that
while we need to capture the same high-level semantic information from
each video, the low-level features corresponding to those semantics
may differ significantly.  The last
fully-connected layer abstracts spatial-domain information from the two
perspectives as two $D$-dimensional feature vectors. To train the network, we
present known true and false correspondences, and use a contrastive
loss function that minimizes sum-of-squares between feature vectors of
true pairs and a hinge loss that examines if  the
distance is greater than a margin for negative pairs (detailed below).

\subsubsection{Motion-domain network}

Figure~\ref{fig:nets}(b) shows the motion-domain network, which learns
correspondences between motion in a \fp video \textit{induced by}
the camera wearer's movements and their  \textit{directly visible} movements
in a \tp video. The idea is that (1) body movements of the camera
wearer (e.g., walking) will be reflected in both \fandtp videos
and that (2) hand motion may also be captured in both cameras during
gestures or actions (e.g., drinking coffee).  We first compute
optical flow for each video, and then stack the flow fields for sets
of five consecutive frames as input to the network.  The \fp
input is the entire optical flow field, whereas 
the \tp input is the flow field \textit{cropped around a single person.} This differs
from the input to the spatial-domain network:
here we
encourage correspondences between the motion of a
camera wearer and their motion as seen by a \tp camera, whereas with
the spatial network we encouraged correspondences
between the \fp scene and the \tp scene 
\textit{except for} the camera wearer.

\subsubsection{Two-stream networks}
\label{sec:twostream}

To combine evidence from both spatial- and motion-domain features, we
use a two-stream network,  as shown in
Figure~\ref{fig:nets}(c).  
Like the spatial-domain network described
above, the spatial stream receives pairs of corresponding \fp and
masked \tp frames, while the temporal stream receives pairs of
corresponding \fp and cropped \tp stacked flow fields. Within each
stream, the final two convolution layers and fully-connected layers
are shared, and then two final fully-connected layers and a
contrastive loss combine the two streams. This design was inspired by
Simonyan and Zisserman~\cite{simonyan2014two},
although that network was proposed for a completely different problem
(activity recognition with a single  static camera)
and so was significantly simpler, taking a single frame and
corresponding stack of optical flow fields. In contrast, our network
features two semi-Siamese streams, two shared fully connected layers,
and a final contrastive loss.

%% file: loss_function.tex
\subsection{Loss functions}\label{sect:lossfunction}

We propose two loss functions for 
learning the distance metric: a standard contrastive loss,
and a new 
``triplet'' loss that considers pairs of correct and incorrect matches. 

\paragraph{Contrastive loss: }
For the Siamese or semi-Siamese networks, we want \fandtp frame representations generated by the CNN
to be close only if they correspond to the same person.
For a batch of $B$ training exemplars, let $x_e^i$ be the \fp visual feature corresponding to the $i$-th exemplar,
$x_p^i$ refer to the \tp visual feature of the $i$-th exemplar, and $y^i$ be an indicator  that
is 1 if the exemplar is a correct correspondence and 0 otherwise.
We define the contrastive loss to be a Euclidean distance for positive exemplars and a hinge loss for negative ones,

\begin{equation} 
\begin{split}
L_{siam}(\theta) =& \sum_i^B y_i ||x_e^i - x_p^i||^2 + \\[-8pt]
          & \relphantom{\sum} (1-y_i) \max(m-||x_e^i - x_p^i||, 0)^2
\end{split}
\end{equation}
where $m$ is a predefined constant margin.


\paragraph{Triplet loss: }

At training time,
 given a \tp video with multiple people and a \fp video, 
we know which pairing is correct  and which pairings are
 not. As an alternative to treating pairs independently as with the
 contrastive loss, we propose to form triplet exemplars consisting
 of both a positive and negative match.  
%
%
%
%
%
%
The triplet loss encourages
 a metric such that the distance from the \fp frame to the
correct \tp frame is low, but to the incorrect \tp frame is high.
More precisely,
for a batch of $B$ training examples,  the $i$-th exemplar is a triple
$(x_e^i, x_1^i, x_0^i)$ corresponding to the features of the \fp frame, the
correct \tp frame, and the incorrectly-masked \tp frame, respectively.
Each exemplar thus has a positive pair $(x_e^i, x_1^i)$ and a negative pair
$(x_e^i, x_0^i)$, and we want to minimize the distance between the true
pair while ensuring the distance between the false pair is  larger.
We use a hinge loss to penalize
if this condition is violated,
\begin{equation} \label{eq:triplet_loss}
\begin{split}
L_{trip} =& \sum_i^B ||x_e^i - x_1^i||^2 +\\[-8pt]
            &\relphantom{\sum}\max(0,m^2-( ||x_e^i - x_0^i||^2 - ||x_e^i - x_1^i||^2 ))
\end{split}
\end{equation}
where $m$ is a constant. This loss  is similar to the
Siamese contrastive loss function, but explicitly enforces the
distance difference to be larger than a margin.  Our loss can be
viewed as a hybrid between Schroff \etal~\cite{schroff2015facenet} and
Bell and Bala~\cite{bell15siggraph}: like~\cite{bell15siggraph}, we
explicitly minimize the distance between the positive pair, and
like~\cite{schroff2015facenet}, we maximize the \textit{difference in
distance} between the negative and positive pairs.

Figure~\ref{fig:nets}(d) shows
the two-stream semi-Siamese network with a triplet loss function.
During training, the spatial stream of the
network expects a \fp frame, a corresponding masked \tp frame, and an
incorrect masked \tp frame, while 
the temporal stream expects a first- and two \tp
cropped optical flow stacks, with the \tp inputs sharing all layers
and the \fandtp layers separate.


%% file: experiment.tex
\section{Experiments}
\label{sec:exp}

We evaluated our proposed technique to identify
people appearing in \tp video and their corresponding \fp videos,
comparing our various network architectures, feature types, and loss
functions against  baselines.


\input{data}

\subsection{Evaluation and training setting}


We use two different metrics for measuring accuracy on the person
correspondence task.  In the first measure, we formulate the problem
in terms of binary classification, asking whether  a given
person in a \tp frame corresponds with a given \fp frame or not,
and then applying this classifier on all possible pairs in each frame.
In this setting, a given \fp video may not correspond to any of
the people in the \tp frame (if the person is out of the camera's field of view), in which 
case the system should reject all candidate pairs.
In the second measure, we formulate the task as the multi-class
classification problem of assigning a given \fp video to a
corresponding person in the \tp scene.  For instance, if there are
four people appearing in the \tp camera, the goal is to choose the one
corresponding to the \fp video, making this a four-way classification
task.


We implemented our networks in Caffe~\cite{jia2014caffe} 
with stochastic gradient descent with fixed learning rate $10^{-5}$,
momentum $0.9$ and weight decay 0.0005 for 50,000 iterations,
using three NVidia Titan X GPUs.
This required about six hours for the spatial network and one
day for the temporal and two-stream networks. 
We have released data
and code online.\footnote{{\scriptsize{\url{http://vision.soic.indiana.edu/identifying-1st-3rd}}}}
As described above, during training we feed our networks with \fp
frames and flow fields, and corresponding positive and negative
cropped flow fields (for the motion networks) and masked images (for
the spatial networks). During testing, we use our
ground-truth bounding boxes to ``highlight'' a person of
interest in the \tp view by masking them out for the spatial network
and cropping them out for the motion networks.

\begin{figure*}
\begin{center}
\includegraphics[clip, trim=0 0 0 0, width=5.5in ]{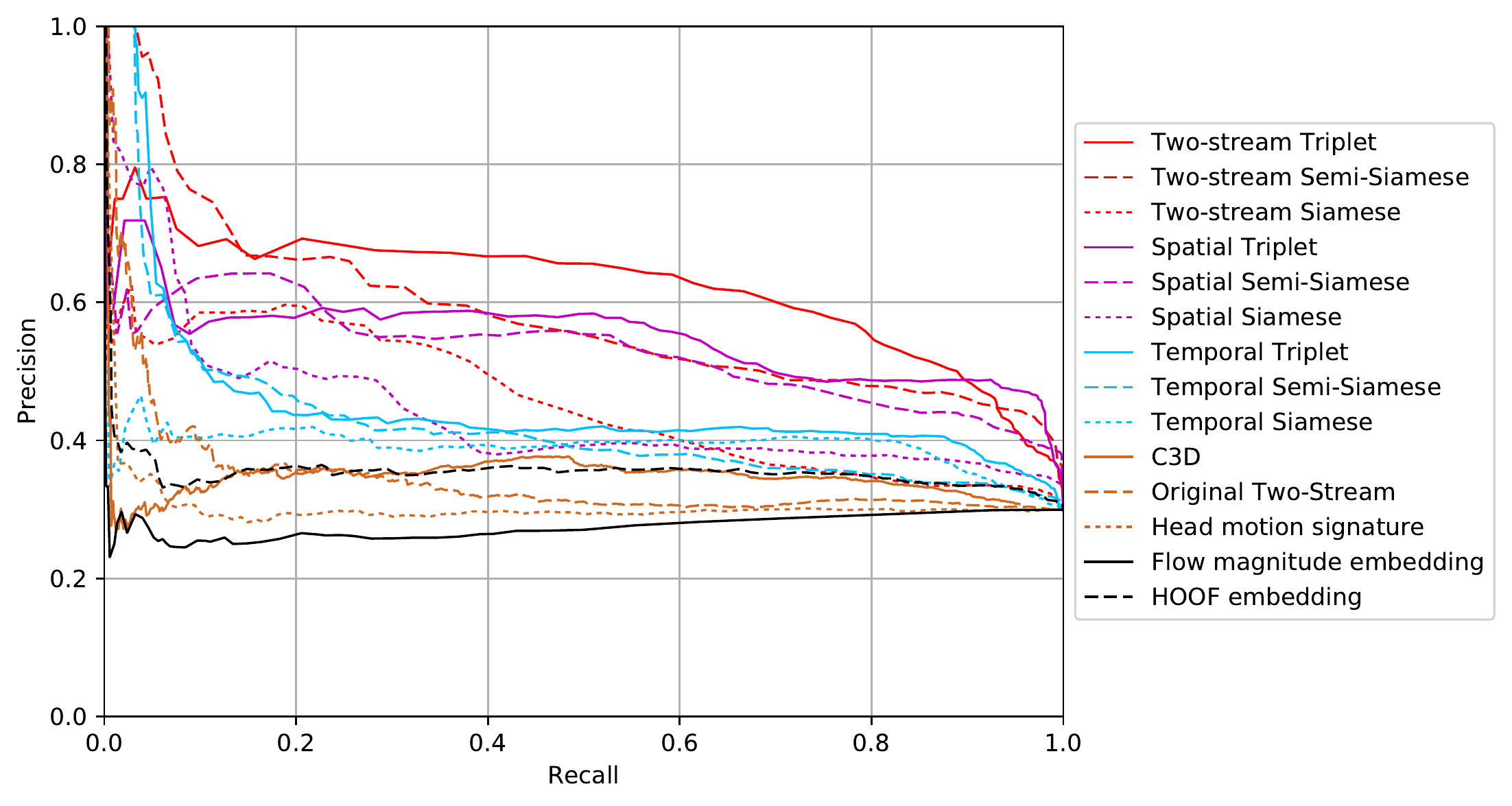}
\end{center}
\vspace{-16pt}
   \caption{\textit{Precision-recall curves} for baselines and variants of our proposed approach.}
\label{fig:pr}
\end{figure*}

\subsection{Baselines}

We implemented multiple baselines to confirm the effectiveness of our
approach. These included mapping of optical flow features from \fp to
\tp view, direct matching of pre-trained CNN features, and learning
an embedding space with traditional HOOF features.


\emph{Flow magnitude to magnitude} calculates the
  mean magnitude of the optical flow vectors on each corresponding
  \fandtp frame, and then learns a linear regressor relating the
  two. Intuitively, at any moment in time there should be a correlation
  between the ``quantity'' of motion in a person's \fp view and
  that of their corresponding appearance in a \tp view.
\emph{HOOF to HOOF} divides the flow field of an image
  into a $3 \times 3$ grid, and then computes 5-bin Histogram of
  Optical Flow (HOOF) features~\cite{hoof} for each cell. We stack
  these 9 histograms to give a 45-d histogram per frame, and then
  average the histograms over a 10-frame temporal window to
  give a final 45-d feature vector. We then learn a linear regressor
  relating the corresponding \fp and \tp HOOFs.
\emph{Odometry to HOOF} estimates camera trajectories
  through visual odometry for each \fp video. We use
  LibVISO2~\cite{Geiger2011IV} to estimate a 13-d pose and velocity
  vector encoding 3-d position, 4-d orientation as a quaternion, and
  angular and linear velocity in each axis for each \fp frame, and
  then learn a regressor to predict the HOOF features in the \tp
  video.
\emph{Velocity to flow magnitude} learns a regressor between
  just the 3-d XYZ velocity vector computed by LibVISO2 for the \tp frame
  and the mean flow magnitude in the \fp frame.

In addition to the above basic baselines, we tested two types of
stronger baselines: (1) directly comparing standard video CNN
features (\emph{two-stream} \cite{simonyan2014two} and \emph{C3D}
\cite{tran2015learning}) from \fandtp videos, and (2) learning an
\emph{embedding} space with traditional HOOF (or motion magnitude). In
particular, the latter baselines have exactly the same loss function
as ours by using fully connected layers.
Finally, we implemented Poleg \textit{et al.}'s  \emph{head
  motion signatures}~\cite{poleg2014head}, which track bounding boxes
of people in \tp frames and correlates them with average XY
optical flows in \fp frames.

\subsection{Results}

Figure~\ref{fig:pr} presents precision-recall curves for variants of
our technique and the baselines, 
and Table~\ref{tab:exp_results} summarizes in terms of Average Precision (AP).  The figure views our task in
terms of retrieval, which is our first measure: for each frame, we generate the set of all
possible candidate pairings consisting of a person in the \tp view and
one of the \fp views, and ask the system to return the correct matches (potentially none).
%
%
The figure shows that for all feature types, our proposed semi-Siamese
architecture outperforms Siamese networks, suggesting that \fandtp
perspectives are different enough that early layers of the CNNs should
be allowed to create specialized low-level features. Switching to the
triplet loss yields a further performance boost compared to the
traditional contrastive loss; for the two-stream network, for example, it increases from
an average precision of 0.585 to 0.621.

Across the different feature types, we find that the spatial-domain
networks perform significantly better than the temporal
(motion)-domain network (e.g., average precisions of 0.549 vs 0.456 for triplet semi-Siamese). The temporal networks still significantly outperform a random baseline (about 0.452 vs 0.354), indicating that motion features contain useful information for matching between views.
The two-stream network that incorporates both types of features yields a further significant improvement (0.621).

Table~\ref{tab:exp_results} 
 clearly
indicates that our approach of learning the shared embedding space for
\fandtp videos significantly outperforms the baselines. Unlike
previous work relying on classic hand-crafted features like head
trajectories (e.g., \cite{poleg2014head}), our method learns the
optimal embedding representation from training data in an end-to-end
fashion, yielding a major increase in accuracy.  We also compared our
Siamese and semi-Siamese architectures against the model of not
sharing any layers between \fp and \tp branches (\textit{Not-Siamese}
in Table~\ref{tab:exp_results}), showing that
semi-Siamese yields better accuracy.


\vspace{-10pt}
\paragraph{Multi-class classification:}

Table~\ref{tab:exp_results} also presents accuracies under our second
evaluation metric, which views the problem as multi-way classification
(with the goal to assign a given \fp video to the correct person in
the \tp scene; e.g., if there are four people in the \tp
video, the goal is to choose the one corresponding to the \fp video).
We see the same pattern as with average precision: semi-Siamese works
better than Siamese, triplet loss outperforms contrastive, the
two-stream networks outperform the single-feature networks, and all of
the baselines underperform. Our proposed two-stream semi-Siamese
network trained with a triplet loss yields the best accuracy, at about
69.3\% correct classification.

\begin{table}
\centering
\ra{1.2}
{\scriptsize{\textsf{
\begin{tabular}{@{}l@{}ll@{}lcc@{}c@{}}\toprule
&\multicolumn{2}{c}{\textit{Network setting}} & & \multicolumn{2}{c}{\textit{Evaluation}} & \phantom{i} \\
\cmidrule{2-3} \cmidrule{5-7} 
&Type &Method && Binary AP  & Multi Accuracy \\ \midrule
&\multirow{9}{*}{Baselines} & Flow magnitude to magnitude &&0.285 &0.250 \\
&&HOOF to HOOF&&0.316 &0.336 \\
&&Odometry to HOOF&&0.302 &0.493 \\
&&Velocity to flow magnitude&&0.279 &0.216 \\
&&HOOF embedding && 0.354 & 0.388 \\
&&Magnitude embedding && 0.276 & 0.216\\
&&Head Motion Signature~\cite{poleg2014head} && 0.300 &0.290 \\
&&Original Two-stream~\cite{simonyan2014two} && 0.350 &0.460 \\
&&C3D~\cite{tran2015learning} && 0.334 &0.505 \\
\midrule
& \multirow{3}{*}{Spatial}& Siamese &&0.481 &0.536 \\
&&Semi-Siamese  &&0.528 &0.585 \\
&&Triplet  &&0.549 &0.588 \\
&\multirow{3}{*}{Temporal} & Siamese &&0.337 &0.372 \\
&&Semi-Siamese  &&0.389 &0.445 \\
&&Triplet&&0.452 &0.490 \\
\midrule
&\multirow{4}{*}{Two-Stream} & Siamese &&0.453 &0.491 \\
&&Not-Siamese && 0.476 & 0.554 \\
&&Semi-Siamese  &&0.585 &0.639 \\
&&Triplet&& \textbf{0.621} & \textbf{0.693} \\
\bottomrule
\end{tabular}
}}}
\caption{Evaluation in terms of average precision and multi-way classification for baselines and variants of our approach.}
\label{tab:exp_results}
\vspace{12pt}
\end{table}

\vspace{-10pt}
\paragraph{Multiple wearable cameras:}

Although we have focused on static \tp cameras,
our approach is applicable to any setting where there
are at least two cameras, one from an actor's viewpoint and another
observing the actor (including multiple wearable cameras).  To test
this, we also tested a scenario in which video from one wearable camera is
treated as \fp while video from the other (wearable) camera is treated
as third-person. These videos seldom have any spatial overlap in their views, and we made our approaches and the baselines to rely only on temporal information for the matching. Table \ref{tab:exp_results_extra} illustrates the results,
showing that our approach outperforms baselines.

\begin{table}[t]
\centering
\ra{1.0}
{\scriptsize{\textsf{
\begin{tabular}{@{}l@{}ll@{}lcc@{}c@{}}\toprule
&\multicolumn{2}{c}{\textit{Network setting}} & & \multicolumn{2}{c}{\textit{Evaluation}} & \phantom{i} \\
\cmidrule{2-3} \cmidrule{5-7} 
&Type &Method && Binary AP  & Multi Accuracy \\ \midrule
&\multirow{8}{*}{Baselines} & Flow magnitude to magnitude &&0.389 &0.442 \\
&&HOOF to HOOF&&0.382 &0.365 \\
&&Odometry to HOOF&&0.181 &0.077 \\
&&Velocity to flow magnitude&&0.310 &0.327 \\
&&HOOF embedding  &&0.405 &0.365 \\
&&Magnitude embedding  &&0.406 &0.442 \\
&&Head Motion Signature~\cite{poleg2014head}  &&0.359 &0.462 \\
&&C3D~\cite{tran2015learning} && 0.380 &0.327 \\
&&Two-stream~\cite{simonyan2014two} (temporal part) && 0.336 & 0.365 \\
\midrule
&\multirow{2}{*}{Ours} &Temporal Semi-Siamese  && \textbf{0.412} & \textbf{0.500} \\
&&Temporal Triplet && 0.386 &\textbf{0.500} \\
\bottomrule
\end{tabular}
}}}
\caption{Results for multiple wearable camera experiments.}
\label{tab:exp_results_extra}
\end{table}

%% file: data.tex
\subsection{Data}
\label{sec:data}

Groups of three to four participants were asked to perform everyday
activities in six indoor environments while two wore \fp video
cameras. Each environment was also equipped with a static camera that
captured third-person video of the room, typically from a perspective
a bit above the participants' heads.  We did not give specific
instructions but simply asked participants to perform everyday,
unstructured activities and interactions, such as shaking hands,
writing on a whiteboard, drinking, chatting, eating, etc. The \fp
videos thus captured not only objects, participants, and background,
but also motion of other people in the scene and ego-motion from hand
and body movements.  Participants were free to walk around the room and
so regularly entered and exited the cameras' fields-of-view.


We collected seven sets of three synchronized videos (two
first- and one third-person) ranging between 5-10
minutes. Three sets had three participants and four included four. All
videos were recorded at HD resolution at 30fps, using Xiaoyi Yi Action
Cameras~\cite{xiaoyi} for the \fp video and a Macbook Pro webcam for
the \tp video.  After collecting the videos, we subsampled to 5fps to
yield 11,225 frames in total.  We created ground truth by manually
drawing bounding boxes around each person in each frame and giving
each box a unique person ID, generating a total of 14,394 bounding
boxes across 4,680 frames.


Because contiguous frames are typically highly correlated, we split
training and test sets at the video level, with five videos for training (3,622 frames) and two for testing (1,058 frames).
Since there are usually
multiple people  per \tp frame, most frames generate
multiple examples of correct and incorrect person pairs (totaling
3,489 positive and 7,399 negative pairs for training, and 1,051
positive and 2,455 negative pairs for testing).
Training and test sets have videos of different scenes and actors.

%% file: case_study.tex
\subsection{Discussion}

\textbf{Generality:} Our approach is 
designed not to rely on long-term tracking and is thus suitable for crowded scenes.
Our
matching is applicable as long as we have a short tracklet of the
corresponding person detected in the \tp video (e.g., only 1
frame in our spatial network), to check whether the match score is
above the threshold.

\begin{figure}
\begin{center}
{\small{
\begin{tabular}{cc}
\includegraphics[clip, trim=100 200 650 50, width=1.5in ]{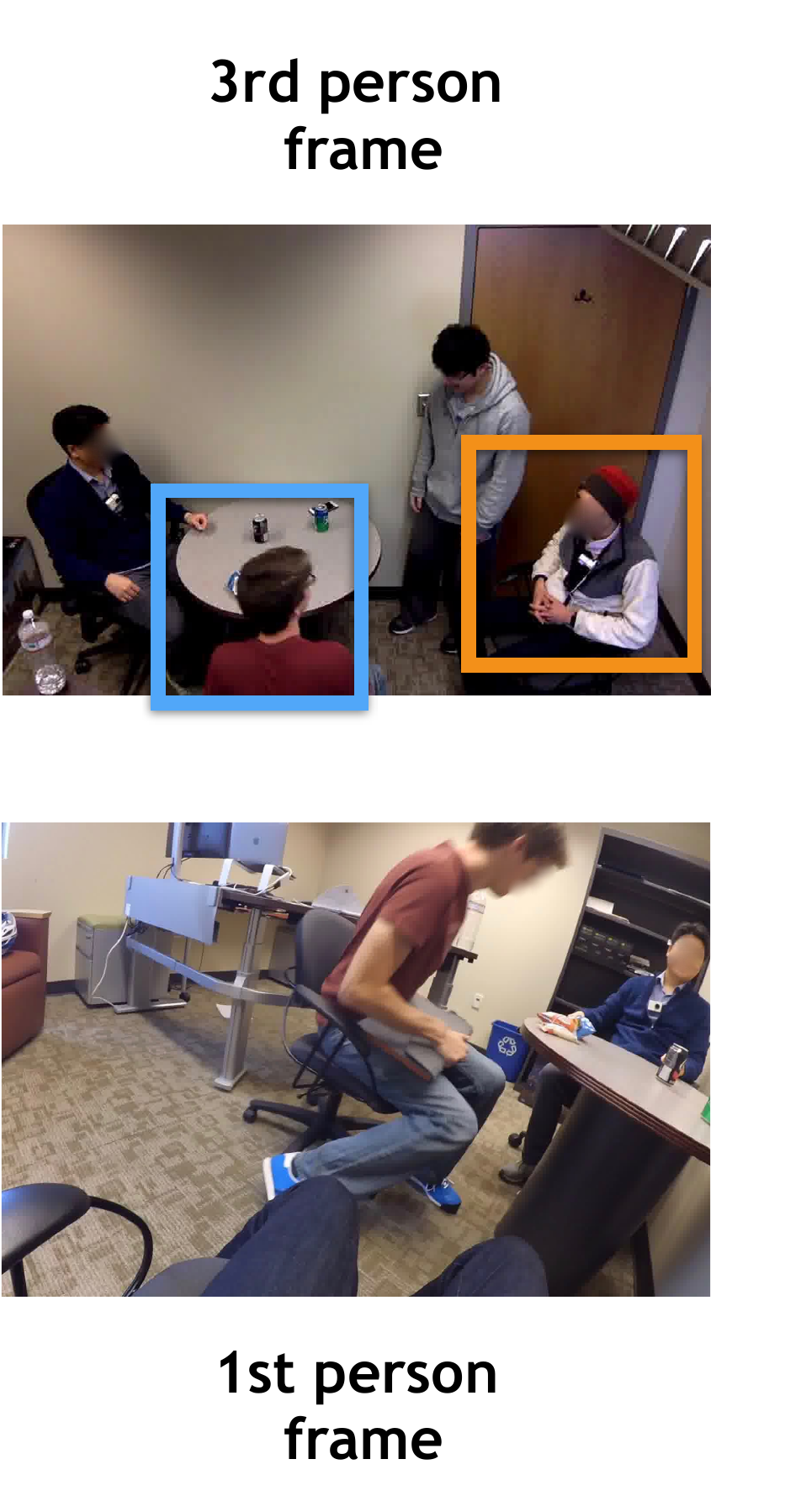} &
\includegraphics[clip, trim=350 200 400 50, width=1.5in ]{image/cases.pdf}  \\
(a) Motion failure case &
(b) Spatial failure case \\
\end{tabular}
}}
\end{center}
\vspace{-6pt}
   \caption{\textit{Sample failures,} with
     the person whose camera took the bottom frame in orange and our incorrect estimate in blue.}
\label{fig:case}
\end{figure}

\textbf{Failure cases:} We observed two typical failure cases.  The
first arises when the actual \fp camera wearer happens to have very
similar motion to another person in the \tp
video. Figure~\ref{fig:case}(a) shows such a situation. Our analysis
of optical flows of the people suggests that the person in blue was in
the process of sitting down, while the camera wearer in orange was
nodding his head, creating confusingly similar flow fields (strong
magnitudes in the vertical direction). Another common failure occurs
when the camera wearer is heavily occluded by another person in the
\tp video, such as in Figure~\ref{fig:case}(b).



\textbf{Gaze:} In addition to our approach of presenting the spatial-domain network 
with 
person regions masked out, we also tried explicitly
estimating gaze of people appearing in \tp videos. The idea was to
encourage the spatial network to focus on the region a person is
looking at, and then match it with \fp videos. We tried Recasens \textit{et
al.}~\cite{recasens-nips2015} for gaze
estimation, but this provided noisy estimates which 
harmed the matching ability of our network.

%% file: conclusion.tex
\section{Conclusion}
We presented a new Convolutional Neural Network framework
to learn distance metrics between \fandtp videos. 
We
found that a combination of three innovations achieved the best results:
(1) a semi-Siamese structure, which takes into account
different features of \fandtp videos (as opposed to full Siamese), (2) a two-stream CNN structure which combines
spatial and motion cues (as opposed to a single stream), and (3) a
triplet loss which explicitly enlarges the margin between
\fandtp videos (as opposed to Siamese contrastive loss).
We hope this paper inspires more work in this important problem of finding correspondences between multiple \fandtp cameras.
